\documentclass{article}

\usepackage{microtype}
\usepackage{graphicx}
\usepackage{subcaption}
\usepackage{booktabs}
\usepackage{multirow}

\usepackage{hyperref}

\usepackage{listings}
\usepackage{xcolor}

\usepackage[preprint]{neurips_2026}

\usepackage{amsmath}
\usepackage{amssymb}
\usepackage{mathtools}
\usepackage{amsthm}

\usepackage[capitalize,noabbrev]{cleveref}
\crefname{lstlisting}{listing}{listings}
\Crefname{lstlisting}{Listing}{Listings}

\theoremstyle{plain}

\theoremstyle{definition}

\theoremstyle{remark}

\usepackage[textsize=tiny]{todonotes}

\title{Your Transformer Can Hold Two Thoughts at Once: Evidence of Linear Superposition in LLMs}

\author{
  Pavel Tikhonov \\
  \And
  Anton Korznikov \\
  \And
  Matvey Mikhalchuk \\
  \AND
  Nikita Dragunov \\
  \And
  Temurbek Rahmatullaev \\
  \And
  Polina Druzhinina \\
  \AND
  Anton Razzhigaev \\
  \And
  Ivan Oseledets \\
  \And
  Elena Tutubalina \\
}

\begin{document}

\maketitle

\begin{abstract}

While Large Language Models (LLMs) rely on highly non-linear components, in this work we demonstrate that they exhibit fundamental linearity: when inputs from distinct text streams are linearly combined, the model outputs a superposition of the individual next-token distributions. We term this the \textit{Superposition Linearity Hypothesis}. We provide evidence that superposition is an intrinsic property of the Transformer architecture rather than an emergent consequence of training; in fact, we observe that it tends to diminish as pretraining progresses. However, we demonstrate that linearity can be substantially restored through lightweight fine-tuning, significantly reducing the divergence between the predicted next-token distribution and the average of the individual next-token distributions. Finally, we introduce a guided decoding procedure that disentangles superposed outputs, enabling the simultaneous generation of two coherent continuations from a single forward pass.
\end{abstract}

\section{Introduction}

The Transformer architecture~\cite{vaswani2017attention} underlies modern Large Language Models (LLMs) and is built from highly non-linear components, including self-attention and MLP blocks with non-linear activations. The prevailing paradigm therefore treats inference as a single coherent semantic stream: to process multiple independent streams one typically runs separate forward passes, performs sequential processing, or modifies the architecture to avoid destructive interference between inputs.

At the same time, recent work shows that, despite these non-linearities, decoder-only Transformers exhibit strong linear structure in the residual stream: transitions between consecutive layers can often be well-approximated by affine maps~\cite{TransformerIsSecretlyLinear}. This motivates a natural question: does such linearity extend beyond layer-to-layer geometry to the model's end-to-end input--output behavior? Specifically, are the computations sufficiently linear that the response to a linear combination of inputs approximates a corresponding combination of their independent outputs?

\begin{figure}[!t]
    \centering
    \includegraphics[width=0.5\linewidth]{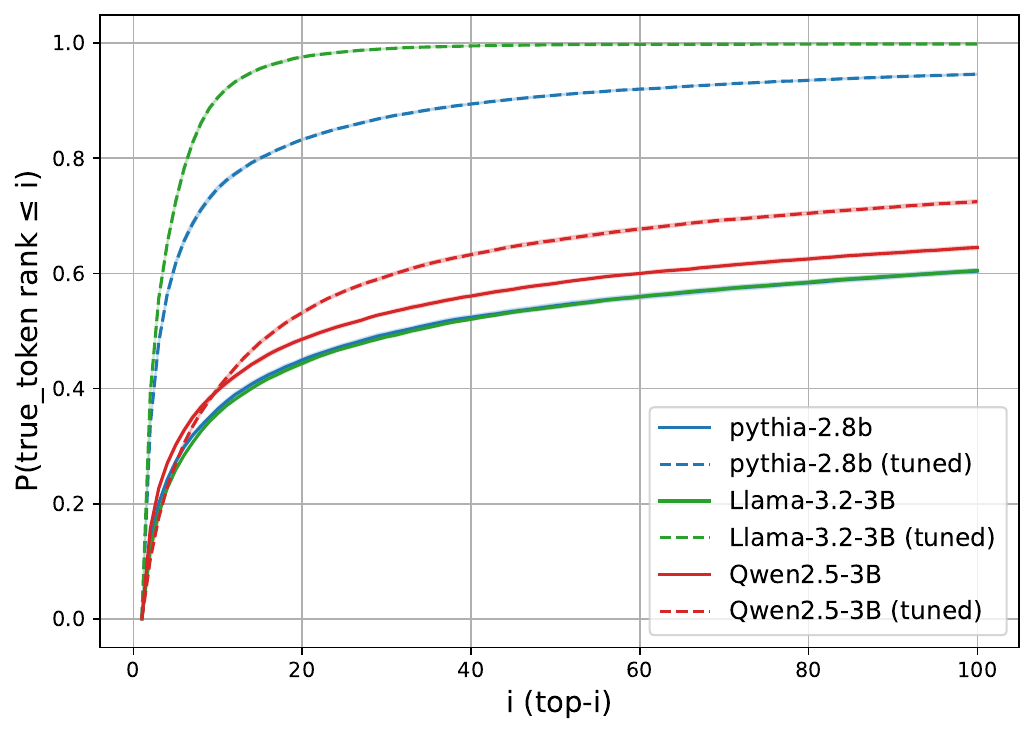}
    \caption{\textbf{Intrinsic superposition in next-token ranks.}
    We mix two prefixes $A$ and $B$ by token-wise averaging embeddings, and obtain mixed logits \(\ell_{\text{mix}}\).
    We then take the single-stream next-token prediction \(\hat{t}=\arg\max \ell_A\) (and symmetrically for \(B\)) and measure its rank under \(\ell_{\text{mix}}\).
    The plot shows \(P(\mathrm{rank}_{\ell_{\text{mix}}}(\hat{t})\le i)\): the chance that a single-stream predicted token remains in the top-\(i\) of the mixed distribution.
    This probability is already high for top-10 without finetuning and increases markedly after lightweight finetuning.
    }
    \label{fig:rank_curve}
\end{figure}

We formalize this as the \textit{Superposition Linearity Hypothesis}: when two token streams with embeddings $\mathbf{x}_{A}$ and $\mathbf{x}_{B}$ are linearly combined (here, via element-wise averaging), the model processes the mixture as a superposition of the two pathways. Empirically, when embeddings from two distinct documents are averaged token-wise and passed through standard pre-trained LLMs, the next-token predictions associated with both streams consistently retain substantial mass in the mixed distribution; in particular, the ground-truth next tokens for both streams frequently appear within the top-10 ranks of the combined output distribution (Fig.~\ref{fig:rank_curve}).

To distinguish architectural bias from learned capability, we track this phenomenon across the pre-training trajectory. We find that superposition fidelity is maximized at initialization and gradually diminishes as the model optimizes the language modeling objective, indicating that linear superposition is intrinsic to the architecture rather than a capability acquired through learning. We further observe a strong correlation between geometric linearity in hidden states (measured via feature additivity) and rank preservation under superposition.

Although pre-training degrades this property, we show it can be substantially restored via a lightweight fine-tuning phase using less than $0.025\%$ of the original pre-training dataset size. Leveraging the amplified linearity, we then propose a decoding procedure that \textit{disentangles} the mixed hidden state, enabling recovery of the distinct continuations corresponding to the original input texts from a single mixed forward pass.

Our contributions are summarized as follows:
\begin{itemize}
    \item We demonstrate that standard pre-trained LLMs retain high probability mass on the same tokens favored by the respective independent distributions.
    \item We show that superposition linearity is an intrinsic architectural property and tends to degrade during pre-training, rather than a capability acquired through the learning process.
    \item We show that the degraded linearity can be substantially recovered using minimal fine-tuning.
    \item We develop a decoding mechanism to disentangle the mixed output distribution back into its constituent text streams.
\end{itemize}

\section{Intrinsic Linearity in Large Language Models}
\label{sec:intrinsic_linearity}

In this section, we investigate the extent to which standard Transformers process superposed inputs without architectural modifications.

\subsection{Problem Formulation}
\label{subsec:formulation}

We consider a decoder-only Transformer language model $M$, mapping a sequence of tokens from vocabulary $\mathcal{V}$ to a sequence of probability distributions over $\mathcal{V}$. Let $E: \mathcal{V} \to \mathbb{R}^d$ be the token embedding function. For a given input sequence $s = (s_1, \dots, s_T)$, the input representation at position $t$ is typically $h_t^{(0)} = E(s_t)$.

We investigate the model's behavior when processing a superposition of two distinct input sequences, $x$ and $y$, of length $T$. We define the \textit{mixed input embedding} $z$ at position $t$ as the element-wise average of the constituent embeddings:
\begin{equation}
    e_t(z) = \frac{1}{2}\left(E(x_t) + E(y_t)\right).
    \label{eq:superposition}
\end{equation}
This mixed representation $z$ is passed through the frozen pre-trained backbone $M$. The model processes this mixture using standard causal self-attention, where the attention mask allows attending to all prior mixed positions $z_{<t}$. We denote the output logits of the model given the mixed input as $\ell(z) \in \mathbb{R}^{|\mathcal{V}|}$ and the resulting probability distribution as $P_{mix}(z) = \text{softmax}(\ell(z))$.

We hypothesize that previously shown approximate linearity in layer-to-layer transitions extends to the global input-output mapping: specifically, that $M$ acts approximately linearly with respect to input superposition. Formally, we test whether $P_{mix}$ approximates $P_{avg} = 0.5(P(x|A) + P(x|B))$, and whether the ground-truth next tokens for both streams retain high probability mass in $P_{mix}$.

To test this hypothesis, we first conducted an evaluation on unmodified pre-trained models. We evaluated models from the Pythia~\cite{biderman2023pythia}, Qwen~\cite{yang2025qwen3}, Llama~\cite{grattafiori2024llama}, and OLMo~\cite{groeneveld2024olmo}, Gemma~\cite{team2024gemma},
families. For data, we used TinyStories~\cite{eldan2023tinystories} (simplified grammar) and FineWeb~\cite{penedo2024fineweb} (real-world naturalistic web text). From each dataset, we sampled random text pairs $(x^{(A)}, x^{(B)})$, tokenized them, and truncated to fixed context lengths.

\subsection{Rank Analysis}
\label{subsec:rank_analysis}

To quantify the preservation of information under superposition, we analyze the rank of the ground-truth next token within the output distribution of the mixed state. Let $t_A$ be the token predicted by the model for input sequence $A$ (i.e., $\arg\max P(x|A)$). We compute the rank of $t_A$ within the mixed distribution $P_{mix} = \text{softmax}(M( \frac{1}{2}(E_A + E_B) ))$. Ideally, if the superposition were perfectly linear, the output distribution would approximate $0.5(P_A + P_B)$, placing $t_A$ and $t_B$ at the very top of the ranking (ranks 1 and 2). In a standard non-linear neural network, one might expect the sum of embeddings to result in a representation orthogonal to both original semantics, pushing $t_A$ and $t_B$ into the tail of the distribution (rank $\sim |\mathcal{V}|/2$).

\paragraph{Cumulative Rank Distribution}
We computed the cumulative distribution function (CDF) of the ranks, $P(\text{rank} \le k)$, for unmodified pre-trained models. Fig.~\ref{fig:rank_curve} illustrates these curves for the Pythia-2.8B, Llama-3.2-3B, and Qwen2.5-3B models.

Our results reveal that the Transformer architecture possesses a surprising degree of intrinsic linearity. Despite the destructive interference inherent in averaging high-dimensional feature vectors, the ground-truth tokens survive the mixing process with high frequency. Specifically, across these architectures (represented by solid lines in the figure):
\begin{itemize}
    \item In approximately \textbf{30--40\%} of cases, the true token appears within the top-10 ranks.
    \item In \textbf{50--60\%} of cases, the true token is found within the top-50 ranks.
    \item By the top-100 ranks, the recovery rate reaches upwards of \textbf{60--65\%}.
\end{itemize}
Considering the large vocabulary sizes ($|\mathcal{V}| \ge 50,000$), these results indicate that the ``signal'' from the original inputs is preserved well above the noise floor. The mixed state does not collapse into gibberish; rather, it effectively narrows down the search space to a small neighborhood containing the valid continuations for both constituent contexts.

\begin{table}[h]
    \small
    \centering
    \caption{\textbf{Distributional Approximation Metrics.} Comparison of distances between the model output on mixed inputs vs.\ the target mixture distribution at context lengths $L=32$ and $L=512$. The \textit{Ratio} indicates the improvement over a random baseline. Lower is better.}
    \label{tab:dist_metrics}
    \begin{tabular}{l|cc|cc|cc}
        \toprule
        \multirow{2}{*}{\textbf{Model}} & \multicolumn{2}{c|}{\textbf{KL Divergence}} & \multicolumn{2}{c|}{\textbf{JS Divergence}} & \multicolumn{2}{c}{\textbf{Wasserstein}} \\
         & Value & Ratio & Value & Ratio & Value & Ratio \\
        \midrule
        \multicolumn{7}{l}{\emph{Context length $L=32$}} \\
        Pythia-160M & 0.91 & \textbf{0.35} & 0.23 & \textbf{0.40} & 0.27 & \textbf{0.63} \\
        Pythia-410M & 1.54 & 0.39 & 0.36 & 0.50 & 0.30 & 0.69 \\
        Pythia-2.8B & 1.86 & 0.42 & 0.40 & 0.54 & 0.31 & 0.71 \\
        Llama-3.1-8B  & 2.16 & 0.37 & 0.48 & 0.57 & 0.32 & 0.69 \\
        \midrule
        \multicolumn{7}{l}{\emph{Context length $L=512$}} \\
        Pythia-160M & 0.92 & \textbf{0.31} & 0.24 & \textbf{0.39} & 0.25 & \textbf{0.61} \\
        Pythia-410M & 1.49 & 0.35 & 0.37 & 0.49 & 0.27 & 0.65 \\
        Pythia-2.8B & 1.69 & 0.37 & 0.37 & 0.51 & 0.30 & 0.69 \\
        Llama-3.1-8B  & 1.98 & 0.34 & 0.45 & 0.54 & 0.31 & 0.67 \\

        \bottomrule
    \end{tabular}
\end{table}

\subsection{Distributional Shape Preservation}
\label{subsec:distribution_analysis}

Rank-based metrics show that the correct tokens often remain salient under embedding mixing, but they do not capture whether the \emph{full} next-token distribution behaves like a linear mixture. Under the Superposition Linearity Hypothesis, we expect the mixed-input output $P_{\text{mix}}$ to approximate the arithmetic mean of the two independent distributions:
\begin{equation}
P_{\text{target}}(x|A, B) = \frac{1}{2} \left( P(x|A) + P(x|B) \right)
\end{equation}

We quantify the mismatch $\mathcal{D}\!\left(P_{\text{target}}, P_{\text{mix}}\right)$ using KL and Jensen--Shannon (JS) divergences, and a Wasserstein distance computed on the top-$256$ tokens with cosine distance between token embeddings as the ground metric.\footnote{For numerical stability in KL and JS, we applied temperature smoothing ($\tau=1.5$). The Wasserstein distance was computed on raw probabilities for the top-256 tokens, using the \textbf{cosine distance} between token embeddings as the ground metric to capture semantic proximity.}

To make distances comparable across model families and contexts, we report a normalized \emph{Superposition Approximation Ratio}:
\begin{equation}
\mathcal{R}_{\mathcal{D}}
\;=\;
\frac{\mathbb{E}_{(A,B)}\!\left[\mathcal{D}\!\left(P_{\text{target}} \,\|\, P_{\text{mix}}\right)\right]}
{\mathbb{E}_{(A,B)}\!\left[\mathcal{D}\!\left(P(\cdot\mid A) \,\|\, P(\cdot\mid B)\right)\right]}.
\end{equation}
Values $\mathcal{R}_{\mathcal{D}}<1$ indicate that the mixed-state output is closer to the ideal linear mixture than two unrelated contexts are to each other.

Across standard pre-trained models on FineWeb, Table~\ref{tab:dist_metrics} shows $\mathcal{R}_{\mathcal{D}}<1$ consistently for KL, JS, and Wasserstein distances, indicating that $P_{\text{mix}}$ preserves substantial distributional structure of the target mixture rather than collapsing to an unrelated distribution.

\paragraph{Contextual stability.}
We further verify that this property is not localized to specific positions: the Total Variation Distance between $P(z)$ and $P_{\text{target}}$ is slightly higher for the first $\sim$20 tokens and then stabilizes at a constant level across the context window, indicating that the geometric properties required for linear superposition persist as the context becomes increasingly complex (Appendix~\ref{app:tvd_pos}).

\subsection{Linearity Dynamics During Training}
\label{subsec:training_dynamics}

To distinguish architectural bias from learned capability, we track \emph{hidden-state additivity} across the pre-training trajectory of the Pythia family. For each pair $(A,B)$ we run three forward passes (stream $A$, stream $B$, and mixed input as in Eq.~\eqref{eq:superposition}), mean-center each hidden state by per-layer means, and compute the $\ell_2$ distance between the $\ell_2$-normalized mixed hidden state and the $\ell_2$-normalized sum of the two single-stream hidden states (full definition in Appendix~\ref{app:training_dyn_err}); we report the resulting layer-averaged error $\bar{\mathcal{E}}$ (lower is more linear).

\begin{figure}[!t]
    \centering
    \includegraphics[width=0.62\linewidth]{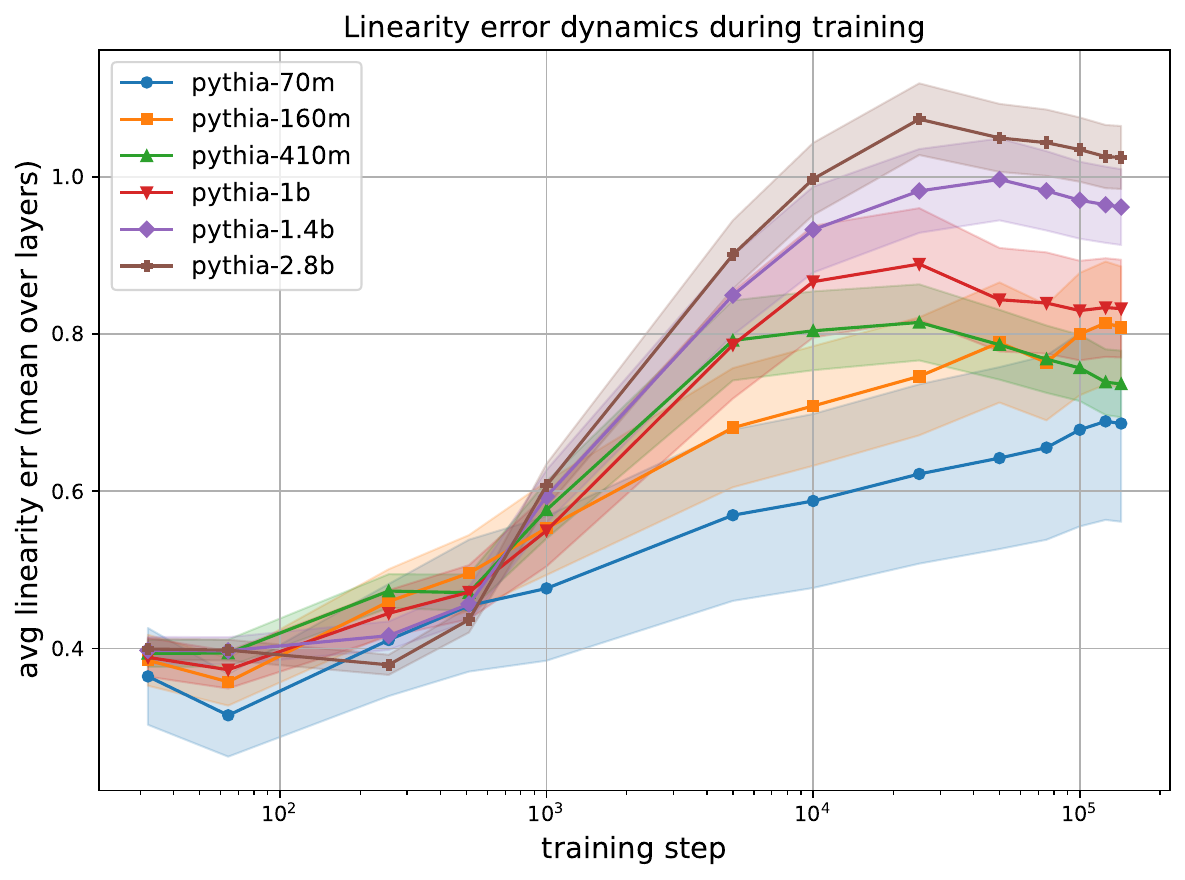}
    \caption{\textbf{Superposition linearity degrades during pre-training.} Mean hidden-state superposition error $\bar{\mathcal{E}}$ (lower is better) across intermediate checkpoints for Pythia models of different sizes; shaded bands show $\pm$ s.e.\ across hidden-state indices.}
    \label{fig:training_dynamics}
\end{figure}

Fig.~\ref{fig:training_dynamics} shows that $\bar{\mathcal{E}}$ is smallest at the earliest checkpoints and grows monotonically as training proceeds, consistent with pre-training amplifying non-linear interactions in the residual stream. A complementary layer-wise linearity analysis~\cite{TransformerIsSecretlyLinear} (Appendix~\ref{app:layerwise}) reveals a U-shaped depth profile in which the deep layers ($\ell \gtrsim 2L/3$) remain near-linear, providing a geometric explanation for why the superposed signal survives through to the output logits.

\paragraph{Scaling beyond two streams.}
To verify that the phenomenon is not specific to the binary case, we extend the rank and distributional analyses of Sec.~\ref{subsec:rank_analysis}--\ref{subsec:distribution_analysis} to $N=3$ by mixing $E(A_t), E(B_t), E(C_t)$ in equal proportions and measuring the ranks of all three ground-truth next tokens. We observe a moderate increase in the approximation ratios (e.g., $\mathcal{R}_{\mathrm{KL}}$ increases by $+0.04$ to $+0.09$ across models; see Appendix~\ref{app:n3_metrics} for full tables). Superposition linearity persists at $N=3$ with quantitative degradation but no qualitative change, indicating the same interference mechanisms operate across stream counts.

\section{An attention-patching analysis}
\label{sec:attention_patching}

The linear superposition demonstrated in Section 2 is counter-intuitive. Key components of the Transformer, particularly self-attention with its softmax non-linearity, are designed to integrate context selectively. One would expect the attention patterns from two unrelated streams (A and B) to interfere destructively, causing the mixed representation to collapse into a state unrelated to either input. Yet, empirically, the signal survives.
This raises the question: does attention play a role in enabling this linearity, or is it a barrier that the residual stream somehow bypasses? To investigate, we design an experiment that disentangles the influence of attention's structural shape from its content-specific computations. We compare our standard embedding mixing setup against two single-stream perturbations: donor patching, which preserves a natural attention structure but decouples it from the text's content, and permutation patching, which destroys the structure while preserving per-token weight distributions.

\paragraph{Setup.}
\label{subsec:patching_setup}
For every text $A$ (FineWeb-Edu, $T{=}128$) we sample an unrelated donor $C$ of the same length and run three forward passes on $A$: (i) a vanilla forward $A_1$; (ii) a \emph{donor-patched} forward $A_d$ in which, at every layer and head, the post-softmax attention weights produced by $C$ are substituted in place of those $A$ would have produced --- the Q/K/V projections, RoPE, and value paths of $A$ are unchanged, only the mixing weights come from $C$; (iii) a \emph{permutation-patched} forward $A_p$ in which, instead of donor weights, we take $A$'s own attention and randomly permute each row within its causal prefix, preserving causality, row-sums, and per-row multisets of weights but destroying positional and content structure. We additionally include a vanilla forward $A'$ on a third unrelated text as the denominator of the Superposition Approximation Ratio. The same FineWeb-Edu pairs and the same content/predictable stratification (detailed in Appendix~\ref{app:patching_examples}) are used throughout. To compare with embedding mixing, we measure the rank of $A_1$'s vanilla top-$1$ token in the perturbed distribution; this is the analogue, for these one-stream perturbations, of the rank metric we used for the two-stream embedding-mixing setup.

\begin{table}[!t]
    \small
    \centering
    \caption{\textbf{Perturbations and fine-tuning on Qwen2.5-3B, stratified by token type.} Median rank of stream $A$'s vanilla $\text{top-}1$ token in the perturbed distribution, and exact-agreement ($\text{top-}1$) percentage. Predictable positions ($\sim\!65\%$ of positions) keep the vanilla $\text{top-}1$ near the top under mixing and donor patching (which preserve attention shape); content positions degrade under base-model mixing and donor patching. Permutation destroys attention shape and collapses across both token types. However, fine-tuning explicitly restores parallel processing on hard content tokens.}
    \label{tab:perturbations_stratified}
    {\setlength{\tabcolsep}{3.5pt}\renewcommand{\arraystretch}{1.05}%
    \begin{tabular}{lcccc}
        \toprule
        & \multicolumn{2}{c}{\textbf{Predictable} (\textasciitilde$65\%$)} & \multicolumn{2}{c}{\textbf{Content} (\textasciitilde$35\%$)} \\
        \cmidrule(lr){2-3}\cmidrule(lr){4-5}
        \textbf{Setup}        & \textbf{med.\ rank} & \textbf{$\text{top-}1$\,\%} & \textbf{med.\ rank} & \textbf{$\text{top-}1$\,\%} \\
        \midrule
        Embedding mixing (Base)       & $6$         & $24.8$ & $284$        & $4.2$ \\
        Donor attention patch (Base)  & $3$         & $33.0$ & $111$        & $5.4$ \\
        Permutation patch (Base)      & $3{,}079$   & $1.3$  & $19{,}246$   & $0.0$ \\
        Embedding mixing (Fine-tuned) & $8$         & $21.6$ & $5$          & $22.8$ \\
        \bottomrule
    \end{tabular}}
\end{table}

\begin{table}[!t]
    \small
    \centering
    \begin{minipage}[t]{.48\textwidth}
        \centering
        \caption{\textbf{Metrics for single-stream perturbations on Qwen2.5-3B.} Permutation destroys attention shape, showing the frequency prior alone is insufficient.}
        \label{tab:attn_patching_single_stream}
        {\setlength{\tabcolsep}{4pt}\renewcommand{\arraystretch}{1.05}%
        \resizebox{\linewidth}{!}{%
        \begin{tabular}{lcc}
            \toprule
            \textbf{Metric}                      & \textbf{Donor patch} & \textbf{Permutation} \\
            \midrule
            median rank $A_1$ $\text{top-}1$     & $8$    & $8{,}148$ \\
            $\text{top-}10$ rank CDF             & $53\%$ & $10.1\%$ \\
            $\mathbb{E}\,\mathrm{KL}$ to vanilla & $3.54$ & $9.13$ \\
            $R_{\mathrm{KL}}$ (vs random)        & $0.27$ & $0.68$ \\
            \bottomrule
        \end{tabular}}}
    \end{minipage}\hspace{0.02\textwidth}
    \begin{minipage}[t]{.48\textwidth}
        \centering
        \caption{\textbf{Embedding mixing vs.\ donor patching.} Mixing retains more signal on hard content prediction.}
        \label{tab:attn_patching_lambada}
        {\setlength{\tabcolsep}{4pt}\renewcommand{\arraystretch}{1.05}%
        \resizebox{\linewidth}{!}{%
        \begin{tabular}{lcc}
            \toprule
            \textbf{Metric}                       & \textbf{Mixing} & \textbf{Donor patch} \\
            \midrule
            median rank $A_1$ $\text{top-}1$      & $19$              & $8$       \\
            \emph{LAMBADA} acc.\ (argmax)         & $\mathbf{2.25\%}$ & $0.5\%$   \\
            \emph{LAMBADA} target med.\ rank      & $339$             & $2{,}350$ \\
            \midrule
            \multicolumn{3}{c}{\emph{LAMBADA baselines~\cite{paperno2016lambada}}} \\
            \multicolumn{3}{c}{LSTM: $0.0\%$, N-Gram: $0.1\%$} \\
            \bottomrule
        \end{tabular}}}
    \end{minipage}
\end{table}

\paragraph{Predictable positions are robust if attention shape is preserved; content positions are not.}
Table~\ref{tab:perturbations_stratified} compares the Qwen2.5-3B forward pass under embedding mixing and the two attention perturbations, broken down by token type. Across predictable positions, the vanilla $\text{top-}1$ token survives with a median rank of $3$--$6$ under embedding mixing and donor patching, and exact agreement remains near $25$--$33\%$. On content positions, however, these two setups diverge: embedding mixing yields a median rank of $284$, while donor patching drops to $111$. Permutation patching, by contrast, collapses entirely across both token types (median rank $3{,}079$ on predictable, $19{,}246$ on content), as it destroys the structural shape of attention. Two implications follow. First, the aggregate recovery numbers we reported in Sec.~\ref{subsec:rank_analysis} are heavily weighted by the predictable majority of positions, where any perturbation that preserves the attention structure (and hence lets the LM-head frequency prior dominate) performs reasonably well. Second, the rank-survival on content positions specifically is what distinguishes the perturbations from each other.

\paragraph{The two attention perturbations dissociate frequency prior from attention shape.}
Read as a self-agreement metric, donor patching looks remarkably benign: median rank $8$ under wholesale substitution of every layer's attention weights, $R_{\mathrm{KL}} = 0.27$ against the unrelated-text baseline. Read as a task-level metric, the same setup is catastrophic: on $200$ LAMBADA prompts (left-truncated to $128$ tokens, target rank measured at the final position), donor patching drives accuracy from the vanilla $73\%$ to $\mathbf{0.5\%}$, with the true target at median rank $2{,}350$. The two read-outs disagree because predictable and content positions disagree --- LAMBADA targets are content words at the final position of long-narrative passages, exactly the regime where the predictable majority does not save us. The permutation control then dissociates two ingredients within donor patching itself. Permutation keeps the LM-head frequency prior intact (it does not touch Q/K/V or the LM head) but destroys the structural shape of attention; this raises $R_{\mathrm{KL}}$ from $0.27$ to $0.68$, drops top-$10$ agreement from $53\%$ to $10\%$, and pushes median rank from $8$ to $8{,}148$. The frequency prior alone is therefore not sufficient to keep aggregate metrics high. What survives donor patching is the joint contribution of two things: the frequency prior (dominating predictable positions) and the structural shape of natural attention --- diagonal/locality bands, attention sinks, head specialization --- properties that natural donor texts share with $A$ even when their content is unrelated. Neither alone is enough.

\paragraph{Embedding mixing carries something beyond ``frequency prior + attention shape''.}
\label{para:mixing_vs_donor}
On the same FineWeb-Edu pairs, embedding mixing has a worse aggregate self-agreement metric than donor patching (median $19$ vs.\ $8$). This is consistent with the model carrying \emph{two} streams' worth of information through the same residual stream rather than one rerouted stream: the natural baseline rank under perfect mixing is $\sim\!1.5$ rather than $1$, and every layer's Q/K/V is contaminated with both inputs from layer $0$, not only the attention weights. Yet on LAMBADA, where the predictable majority does not help, the order is reversed: embedding mixing reaches $2.25\%$ raw argmax accuracy with target median rank $339$, while donor patching reaches $0.5\%$ at median rank $2{,}350$ --- a $\mathbf{4.5\times}$ accuracy gap and a $\mathbf{7\times}$ rank gap. Whatever the mixing setup carries on hard content positions, it is more than what survives donor patching, which is the LM-head prior plus structural attention. This places a non-trivial lower bound on what additive embedding composition has to preserve: it is enough to retain meaningfully more case-specific signal on hard content prediction than a wholesale donor-attention swap, while doing so simultaneously for two unrelated streams.

\paragraph{Fine-tuning restores content-position survival.}
\label{para:patching_finetuned}
As we will show in Section~\ref{sec:improving_linearity}, the model can be fine-tuned to better support superposition. The dichotomy between predictable and content tokens clarifies exactly what this fine-tuning achieves. As seen in Table~\ref{tab:perturbations_stratified}, the base model's aggregate median rank of $19$ under embedding mixing is heavily buoyed by the predictable positions (median rank $6$)---its survival on actual content positions is very poor (median rank $284$). However, after fine-tuning, the aggregate median rank improves to $6$. What is surprising here is how this happens: while predictable positions remain largely unchanged (median rank $8$), the content positions see a massive restoration. Their median rank drops all the way to $5$, and exact $\text{top-}1$ agreement jumps to $22.8\%$. The fine-tuned model is therefore no longer just leaning on the frequency prior; it is genuinely processing the semantic content of both streams in parallel.

\section{Improving Linearity with Finetuning}
\label{sec:improving_linearity}

As demonstrated in Sec.~\ref{sec:intrinsic_linearity}, pre-trained Transformer models exhibit an intrinsic, albeit approximate, ability to process superposed inputs linearly. This property is present despite the standard pre-training objective not explicitly incentivizing such behavior. Here, we investigate whether this architectural capability can be enhanced through targeted optimization. We explore if a lightweight fine-tuning phase can align the model's weights to explicitly support superposition.

We employ a self-distillation framework designed to minimize the discrepancy between the model's output on a mixed input and the mixture of its independent outputs. We initialize a student model $M_{student}$ with pre-trained weights and use a frozen copy of the same model as the teacher $M_{teacher}$. For a pair of distinct text sequences $x^{(A)}$ and $x^{(B)}$, we define the target probability distribution as the arithmetic mean of the teacher's independent predictions:
\begin{equation}
    P_{target} = \frac{1}{2} \left( M_{teacher}(x^{(A)}) + M_{teacher}(x^{(B)}) \right).
\end{equation}
The student model processes the element-wise average of the input embeddings $z = \frac{1}{2}(E(x^{(A)}) + E(x^{(B)}))$. The objective is to minimize the Kullback-Leibler (KL) divergence between the student's output and the target mixture:
\begin{equation}
    \mathcal{L} = D_{KL}\left( P_{target} \parallel M_{student}(z) \right).
\end{equation}
We applied this procedure to the Pythia, Qwen and Llama models using a subset of the FineWeb dataset (approximately 200k steps).

Revisiting the metrics of Sec.~\ref{sec:intrinsic_linearity}, fine-tuning substantially reduces the divergence between the predicted and target distributions: on Pythia-2.8B, the mean KL divergence drops from $1.86$ to $0.27$, and the Superposition Approximation Ratio $\mathcal{R}_{\mathrm{KL}}$ drops from $0.42$ to $0.06$ (Table~\ref{tab:dist_metrics}, last row). At the rank level (Fig.~\ref{fig:rank_curve}), the probability of the true next token appearing in the $\text{top-}5$ rises from $\approx\!30\%$ to $>60\%$. The interference observed in the base model is therefore largely reversible, with both ground-truth streams preserved with high fidelity at the output layer.

As we noted in Table~\ref{tab:perturbations_stratified}, what makes this rank restoration interesting is that it doesn't just boost high-frequency predictable tokens. On the hard content tokens---where the base model effectively collapsed (median rank $284$)---the fine-tuned model manages to recover the signal entirely, bringing the median rank down to $5$ and pushing exact $\text{top-}1$ agreement to $22.8\%$. This confirms that our lightweight optimization isn't taking a shortcut; it explicitly rehabilitates the parallel processing of complex, case-specific semantic content.

We note up front that this restoration is not free: the same objective measurably reduces single-stream next-token-prediction quality (e.g.\ Pythia-2.8B LAMBADA $0.544 \to 0.357$; Qwen2.5-3B $0.602 \to 0.460$), and the layer-wise mechanism behind the improvement, together with full PPL/NLL trade-offs on FineWeb, is reported in Appendix~\ref{app:finetuning_extra}. We return to the gap between distributional fidelity and decoding quality in Sec.~\ref{sec:decoding_obstruction}.

\section{Decoding Two Streams Out of a Mixed Forward Pass}
\label{sec:decoding_obstruction}

Sec.~\ref{sec:improving_linearity} showed that the mixed-input distribution can be brought close to its analytical target. A natural last question is whether one can then \emph{decode} the two streams separately --- which is what would turn the phenomenon into a parallel-inference primitive. We show that this is strictly harder than fitting the mixed distribution, and explain why.

\paragraph{The geometric-mean obstruction.}
While the probability mass of the target tokens is largely restored by fine-tuning, sampling directly from the mixed distribution creates semantically inconsistent sequences, as the model alternates between the tokens of context $A$ and context $B$. In a linear superposition where the logits are approximately averaged, the resulting probabilities scale with the geometric mean of the independent distributions:
\begin{equation}
P'_{\text{target}}(t)
\;\propto\;
\exp\!\bigl(\tfrac{1}{2}(\ell_A(t) + \ell_B(t))\bigr)
\;\propto\;
\sqrt{P_A(t)\,P_B(t)}.
\label{eq:sqrt_target}
\end{equation}
This creates an intrinsic decoding challenge: any token that is highly probable in stream $A$ but highly unlikely in stream $B$ is heavily penalized by the geometric mean, pushing its mixed probability down. Thus, achieving rank $1$ for both streams simultaneously using standard decoding is exceptionally difficult, even when both ground-truth tokens reliably appear in the $\text{top-}5$ (as achieved by fine-tuning). Consequently, to make superposition practically useful for parallel processing, we require a mechanism capable of disentangling the mixed hidden state back into its constituent streams. Overcoming this obstruction fully remains an open problem for future research; however, below we propose a proof-of-concept decoding method as a suggestion for the community (with additional decoding variants, including a parameter-free two-head approach and inference-time logit arithmetic, deferred to Appendix~\ref{app:decoding_extra}).

\paragraph{Joint Contrastive decoding.}
\label{subsec:joint_contrastive}
To exhibit a concrete decoder in this regime, we use the \emph{Joint Contrastive} variant in which a small auxiliary model $M_{\text{small}}$ provides per-stream guidance during fine-tuning. The disentangled logits are
\begin{align}
\tilde\ell^{(A)} &= \ell_{\text{large}}(z) + \alpha\,\ell_{\text{small}}(A) - \beta\,\ell_{\text{small}}(B),
&
\tilde\ell^{(B)} &= \ell_{\text{large}}(z) + \alpha\,\ell_{\text{small}}(B) - \beta\,\ell_{\text{small}}(A),
\end{align}
with $z = \tfrac{1}{2}(E(A) + E(B))$. The scalars $\alpha,\beta$ are initialized to $1$ and trained jointly with the backbone on the symmetric per-stream cross-entropy loss (hyperparameters: Appendix~\ref{app:hparams}).

\begin{table}[t!]
    \small
    \centering
    \caption{\textbf{Joint Contrastive decoding as a proof of concept.} LAMBADA mean accuracy across both streams on a superposed forward pass; Jaccard token-overlap on FineWeb generations as a separation metric (lower is better). Joint Contrastive substantially raises accuracy over raw pretrained mixing, but does not close the gap to the small-model single-stream baseline. Two-Head, Mixed Distillation, gradient-optimized inference, and TinyStories LLM-as-judge are in Appendix~\ref{app:decoding_extra}.}
    \label{tab:quant_metrics}
    {\setlength{\tabcolsep}{3.5pt}\renewcommand{\arraystretch}{1.05}%
    \resizebox{\linewidth}{!}{%
    \begin{tabular}{llcccc}
        \toprule
        \textbf{Backbone} & \textbf{Guide} & \textbf{Method} & \textbf{LAMBADA single} & \textbf{LAMBADA mixed} & \textbf{Jaccard} \\
        & & & (large / small) & (mean acc.) & (FineWeb) \\
        \midrule
        Qwen2.5-3B  & Qwen2.5-0.5B & Pretrained         & $0.592$ / $0.437$ & $0.168$ & $0.126$ \\
        Qwen2.5-3B  & Qwen2.5-0.5B & Joint Contrastive  & $0.592$ / $0.437$ & $\mathbf{0.345}$ & $\mathbf{0.061}$ \\
        \midrule
        Llama-3.2-3B & Llama-3.2-1B & Pretrained        & $0.643$ / $0.540$ & $0.182$ & $0.094$ \\
        Llama-3.2-3B & Llama-3.2-1B & Joint Contrastive & $0.643$ / $0.540$ & $\mathbf{0.430}$ & $\mathbf{0.067}$ \\
        \midrule
        Pythia-2.8B  & Pythia-160m  & Pretrained        & $0.544$ / $0.225$ & $0.065$ & $0.109$ \\
        Pythia-1.4B  & Pythia-160m  & Joint Contrastive & $0.499$ / $0.225$ & $\mathbf{0.110}$ & $\mathbf{0.080}$ \\
        \bottomrule
    \end{tabular}}}
\end{table}

Table~\ref{tab:quant_metrics} reports LAMBADA mean accuracy on the superposed forward pass. Joint Contrastive lifts mean accuracy from the raw-pretrained $0.06$--$0.18$ into the $0.11$--$0.43$ range while keeping inter-stream Jaccard overlap low, and on Llama-3.2-3B reaches $0.43$ vs.\ a single-stream small-model baseline of $0.54$. We treat this as a proof of concept that the superposed signal is exploitable, with the residual gap to single-stream consistent with the obstruction in Eq.~\eqref{eq:sqrt_target} rather than with a deficiency of fine-tuning.

\section{Conclusion}
In this work, we demonstrated that Transformers can process linearly combined text streams as a superposition of independent pathways. We showed that while this intrinsic architectural property degrades during standard pre-training, it can be robustly restored via lightweight fine-tuning, allowing the model to maintain distinct semantic signals within a mixed hidden state.

These findings have profound implications for efficient deployment. By successfully disentangling superposed outputs, our approach enables the generation of two coherent continuations from a single forward pass, theoretically offering a $\mathbf{2\times}$ increase in inference throughput. Furthermore, since multiple streams are compressed into a single vector representation, this paradigm drastically reduces memory consumption, effectively halving the KV-cache footprint per active stream.

\section{Related Work}
\label{sec:related_work}

Decoder-only Transformers exhibit surprisingly linear behavior in the residual stream: consecutive-layer mappings are often well-approximated by affine transforms \citep{TransformerIsSecretlyLinear}. Closely related are methods that \emph{explicitly} multiplex multiple inputs into a single representation and then demultiplex predictions, e.g., DataMUX \citep{murahari2022datamux}, binding/unbinding-based MIMONets \citep{menet2023mimonets}, and RevMUX for efficient LLM batch inference via reversible adapters \citep{xu2024revmux}. At inference time, superposition is also exploited for parallel generation or context processing: Superposed Decoding mixes draft token embeddings to produce multiple continuations in one autoregressive pass \citep{shen2024superposed}, while superposition prompting accelerates RAG by processing multiple document paths within a single forward pass \citep{merth2024superpositionprompting}. Complementary perspectives include task superposition in in-context learning \citep{xiong2024everything} and feature superposition as a representational bottleneck \citep{elhage2022toy}. In contrast to approaches that add mux/demux structure, we isolate an \emph{intrinsic} input--output superposition effect in standard pretrained LLMs under linear embedding mixing, track its degradation during pretraining, and show it can be restored by lightweight finetuning and partially disentangled at decoding time. Concretely, prior multiplexing methods (DataMUX, MIMONets, RevMUX) treat superposition as an engineered capability that must be externally imposed via dedicated layers, VSA-style binding/unbinding keys, or isometry regularization. Our claim is qualitatively different: superposition is intrinsic to standard pretrained Transformers, simple embedding averaging in off-the-shelf LLMs already preserves substantial signal, and our lightweight fine-tuning \emph{restores} a property that pretraining has degraded rather than \emph{creating} a new one.

\section{Limitations}
\label{sec:limitations}

Our study establishes the existence and recoverability of linear superposition in Transformer language models, but several boundaries of this phenomenon remain to be explored:

\begin{itemize}
    \item \textbf{Context length and language:} Our evaluations utilized relatively short-context ($L \le 128$ for analytical experiments, with extensions to $L=512$ in Sec.~\ref{subsec:distribution_analysis}) and predominantly monolingual text corpora. Generalizing this approach to substantially longer contexts or multilingual settings involves more complex embedding geometries, presenting an important avenue for future work.
    
    \item \textbf{Multimodal models:} Our investigation focuses exclusively on text-based language models. We have not tested whether these linear superposition properties extend to multimodal architectures. Mixing embeddings from intrinsically different modalities (e.g., combining text tokens with image patches) poses distinct structural and geometric challenges for the residual stream that remain completely unexplored.
\end{itemize}

\nocite{langley00}

\bibliography{references}

@inproceedings{vaswani2017attention,
  author    = {Ashish Vaswani and Noam Shazeer and Niki Parmar and Jakob Uszkoreit and Llion Jones and Aidan N. Gomez and Lukasz Kaiser and Illia Polosukhin},
  title     = {Attention Is All You Need},
  booktitle = {Advances in Neural Information Processing Systems},
  volume    = {30},
  year      = {2017}
}

@article{team2024gemma,
  title={Gemma 2: Improving open language models at a practical size},
  author={Team, Gemma and Riviere, Morgane and Pathak, Shreya and Sessa, Pier Giuseppe and Hardin, Cassidy and Bhupatiraju, Surya and Hussenot, L{\'e}onard and Mesnard, Thomas and Shahriari, Bobak and Ram{\'e}, Alexandre and others},
  journal={arXiv preprint arXiv:2408.00118},
  year={2024}
}

@inproceedings{TransformerIsSecretlyLinear,
  author    = {Anton Razzhigaev and Matvey Mikhalchuk and Elizaveta Goncharova and Nikolai Gerasimenko and Ivan V. Oseledets and Denis Dimitrov and Andrey Kuznetsov},
  title     = {Your Transformer is Secretly Linear},
  booktitle = {Proceedings of the 62nd Annual Meeting of the Association for Computational Linguistics (Volume 1: Long Papers)},
  pages     = {5376--5384},
  publisher = {Association for Computational Linguistics},
  year      = {2024},
  doi       = {10.18653/v1/2024.acl-long.293},
  url       = {https://doi.org/10.18653/v1/2024.acl-long.293}
}

@article{biderman2023pythia,
  author     = {Stella Biderman and others},
  title      = {{Pythia}: A Suite for Analyzing Large Language Models Across Training and Scaling},
  journal    = {CoRR},
  volume     = {abs/2304.01373},
  year       = {2023},
  url        = {https://arxiv.org/abs/2304.01373},
  eprinttype = {arXiv},
  eprint     = {2304.01373}
}

@article{yang2025qwen3,
  author     = {An Yang and others},
  title      = {{Qwen3} Technical Report},
  journal    = {CoRR},
  volume     = {abs/2505.09388},
  year       = {2025},
  url        = {https://doi.org/10.48550/arXiv.2505.09388},
  doi        = {10.48550/ARXIV.2505.09388},
  eprinttype = {arXiv},
  eprint     = {2505.09388}
}

@article{grattafiori2024llama,
  author     = {Aaron Grattafiori and others},
  title      = {The {Llama} 3 Herd of Models},
  journal    = {CoRR},
  volume     = {abs/2407.21783},
  year       = {2024},
  url        = {https://doi.org/10.48550/arXiv.2407.21783},
  doi        = {10.48550/ARXIV.2407.21783},
  eprinttype = {arXiv},
  eprint     = {2407.21783}
}

@article{groeneveld2024olmo,
  author     = {Dirk Groeneveld and others},
  title      = {{OLMo}: Accelerating the Science of Language Models},
  journal    = {CoRR},
  volume     = {abs/2402.00838},
  year       = {2024},
  url        = {https://doi.org/10.48550/arXiv.2402.00838},
  doi        = {10.48550/ARXIV.2402.00838},
  eprinttype = {arXiv},
  eprint     = {2402.00838}
}

@article{eldan2023tinystories,
  author     = {Ronen Eldan and Yuanzhi Li},
  title      = {{TinyStories}: How Small Can Language Models Be and Still Speak Coherent English?},
  journal    = {CoRR},
  volume     = {abs/2305.07759},
  year       = {2023},
  url        = {https://doi.org/10.48550/arXiv.2305.07759},
  doi        = {10.48550/ARXIV.2305.07759},
  eprinttype = {arXiv},
  eprint     = {2305.07759}
}

@article{penedo2024fineweb,
  author     = {Guilherme Penedo and others},
  title      = {Decanting the Web for the Finest Text Data at Scale},
  journal    = {CoRR},
  volume     = {abs/2406.17557},
  year       = {2024},
  url        = {https://doi.org/10.48550/arXiv.2406.17557},
  doi        = {10.48550/ARXIV.2406.17557},
  eprinttype = {arXiv},
  eprint     = {2406.17557}
}

@article{paperno2016lambada,
  author     = {Denis Paperno and others},
  title      = {The {LAMBADA} Dataset: Word Prediction Requiring a Broad Discourse Context},
  journal    = {CoRR},
  volume     = {abs/1606.06031},
  year       = {2016},
  url        = {https://arxiv.org/abs/1606.06031},
  eprinttype = {arXiv},
  eprint     = {1606.06031}
}

@article{shen2024superposed,
  author     = {Ethan Shen and Alan Fan and Sarah M. Pratt and Jae Sung Park and Matthew Wallingford and Sham M. Kakade and Ari Holtzman and Ranjay Krishna and Ali Farhadi and Aditya Kusupati},
  title      = {Superposed Decoding: Multiple Generations from a Single Autoregressive Inference Pass},
  journal    = {CoRR},
  volume     = {abs/2405.18400},
  year       = {2024},
  url        = {https://doi.org/10.48550/arXiv.2405.18400},
  doi        = {10.48550/ARXIV.2405.18400},
  eprinttype = {arXiv},
  eprint     = {2405.18400}
}

@article{xiong2024everything,
  author     = {Zheyang Xiong and Ziyang Cai and John Cooper and Albert Ge and Vasilis Papageorgiou and Zack Sifakis and Angeliki Giannou and Ziqian Lin and Liu Yang and Saurabh Agarwal and Grigorios G. Chrysos and Samet Oymak and Kangwook Lee and Dimitris Papailiopoulos},
  title      = {Everything Everywhere All at Once: {LLM}s can In-Context Learn Multiple Tasks in Superposition},
  journal    = {CoRR},
  volume     = {abs/2410.05603},
  year       = {2024},
  url        = {https://doi.org/10.48550/arXiv.2410.05603},
  doi        = {10.48550/ARXIV.2410.05603},
  eprinttype = {arXiv},
  eprint     = {2410.05603}
}

@article{murahari2022datamux,
  author     = {Vishvak Murahari and Carlos E. Jimenez and Runzhe Yang and Karthik Narasimhan},
  title      = {{DataMUX}: Data Multiplexing for Neural Networks},
  journal    = {CoRR},
  volume     = {abs/2202.09318},
  year       = {2022},
  url        = {https://arxiv.org/abs/2202.09318},
  eprinttype = {arXiv},
  eprint     = {2202.09318}
}

@article{menet2023mimonets,
  author     = {Nicolas Menet and Michael Hersche and Geethan Karunaratne and Luca Benini and Abu Sebastian and Abbas Rahimi},
  title      = {{MIMONets}: Multiple-Input-Multiple-Output Neural Networks Exploiting Computation in Superposition},
  journal    = {CoRR},
  volume     = {abs/2312.02829},
  year       = {2023},
  url        = {https://doi.org/10.48550/arXiv.2312.02829},
  doi        = {10.48550/ARXIV.2312.02829},
  eprinttype = {arXiv},
  eprint     = {2312.02829}
}

@inproceedings{xu2024revmux,
  author    = {Yige Xu and Xu Guo and Zhiwei Zeng and Chunyan Miao},
  title     = {{RevMUX}: Data Multiplexing with Reversible Adapters for Efficient {LLM} Batch Inference},
  booktitle = {Proceedings of the 2024 Conference on Empirical Methods in Natural Language Processing},
  pages     = {22072--22087},
  publisher = {Association for Computational Linguistics},
  year      = {2024},
  doi       = {10.18653/v1/2024.emnlp-main.1232},
  url       = {https://aclanthology.org/2024.emnlp-main.1232/}
}

@inproceedings{merth2024superpositionprompting,
  author    = {Thomas Merth and Qichen Fu and Mohammad Rastegari and Mahyar Najibi},
  title     = {Superposition Prompting: Improving and Accelerating Retrieval-Augmented Generation},
  booktitle = {Proceedings of the 41st International Conference on Machine Learning},
  series    = {Proceedings of Machine Learning Research},
  volume    = {235},
  pages     = {35507--35527},
  publisher = {PMLR},
  year      = {2024},
  url       = {https://proceedings.mlr.press/v235/merth24a.html}
}

@misc{elhage2022toy,
  author = {Nelson Elhage and others},
  title  = {Toy Models of Superposition},
  year   = {2022},
  note   = {Technical report}
}

@inproceedings{langley00,
  author    = {Pat Langley},
  title     = {Crafting Papers on Machine Learning},
  booktitle = {Proceedings of the 17th International Conference on Machine Learning},
  year      = {2000}
}
\bibliographystyle{plainnat}

\newpage
\appendix

\section{Compute Resources}
\label{app:compute_resources}

\paragraph{Training cost.}
\label{para:training_cost}
While our distillation objective is lightweight compared to full pre-training, fine-tuning large language models remains computationally intensive. A single fine-tuning run of Pythia-2.8B for $150{,}000$ steps required approximately $114$ hours on $2\times$A100 80GB GPUs. The Qwen2.5-3B distillation run took $132$ hours on the same hardware setup, while Llama-3.2-3B completed in $128$ hours. The Joint Contrastive guided decoding fine-tuning for Pythia-1.4B (with a 160M guide model) required $68$ hours on a single A100 80GB GPU.

\section{Frequency-Baseline Control}
\label{app:frequency_control}

Using single-stream Pythia-2.8B on unrelated FineWeb pairs $(A,B)$, we compute the logits of stream $A$ alone and measure the rank of stream $B$'s ground-truth next token at the same position. This preserves the natural token-frequency distribution while breaking the contextual link between the two streams. Under this control, the rank of the cross-stream ground-truth token is in the top-$3$ in only $1.12\%$ of cases, in the top-$10$ in $2.63\%$, and in the top-$100$ in $10.41\%$. Same-position token overlap between unrelated streams is $0.2\%$. Under the actual superposition forward pass, top-$10$ recovery is $30$--$40\%$ and top-$100$ recovery is $60$--$65\%$, exceeding the frequency-only baseline by an order of magnitude. The effect is therefore not attributable to vocabulary under-utilization or Zipf-like priors.

\section{Hidden-state additivity error}
\label{app:training_dyn_err}

Let $h^{(A)}_{l,t}$, $h^{(B)}_{l,t}$, and $h^{(\text{mix})}_{l,t}$ denote the hidden states at index $l$ and position $t$ under the three forward passes (stream $A$, stream $B$, mixed input as in Eq.~\eqref{eq:superposition}). Mean-centering by per-layer means $\mu^{(\cdot)}_l$ over the evaluation set yields $\tilde h^{(\cdot)}_{l,t} = h^{(\cdot)}_{l,t} - \mu^{(\cdot)}_l$. The hidden-state superposition error is
\begin{equation}
\epsilon_{l,t}(A,B) \;=\;
\left\|
\frac{\tilde h^{(\text{mix})}_{l,t}}{\|\tilde h^{(\text{mix})}_{l,t}\|_2}
-
\frac{\tilde h^{(A)}_{l,t} + \tilde h^{(B)}_{l,t}}{\|\tilde h^{(A)}_{l,t} + \tilde h^{(B)}_{l,t}\|_2}
\right\|_2,
\label{eq:train_dyn_err}
\end{equation}
and we report $\mathcal{E}_l = \mathbb{E}_{(A,B),t}[\epsilon_{l,t}]$ together with its layer-averaged summary $\bar{\mathcal{E}}$ (lower indicates stronger superposition linearity).

\section{Confidence-vs-Rank Curve}
\label{app:confidence}

\begin{figure}[h]
    \centering
    \includegraphics[width=0.7\linewidth]{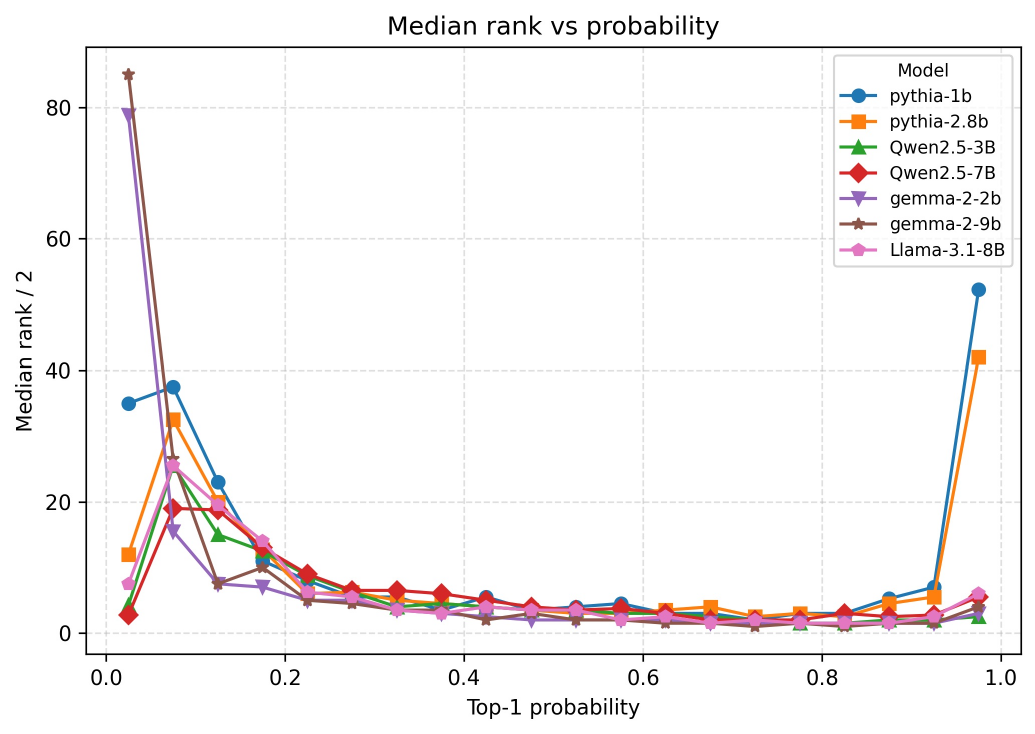}
    \caption{\textbf{Robustness of confident predictions.} Median rank of the true token in the mixed distribution as a function of its probability in the original independent forward pass. Tokens predicted with high confidence ($P > 0.5$) almost always survive the superposition process (median rank $\approx 3$), whereas low-confidence predictions are more susceptible to interference.}
    \label{fig:rank_vs_prob}
\end{figure}

\section{Contextual Stability of Superposition}
\label{app:tvd_pos}

\begin{figure}[h]
    \centering
    \includegraphics[width=0.7\linewidth]{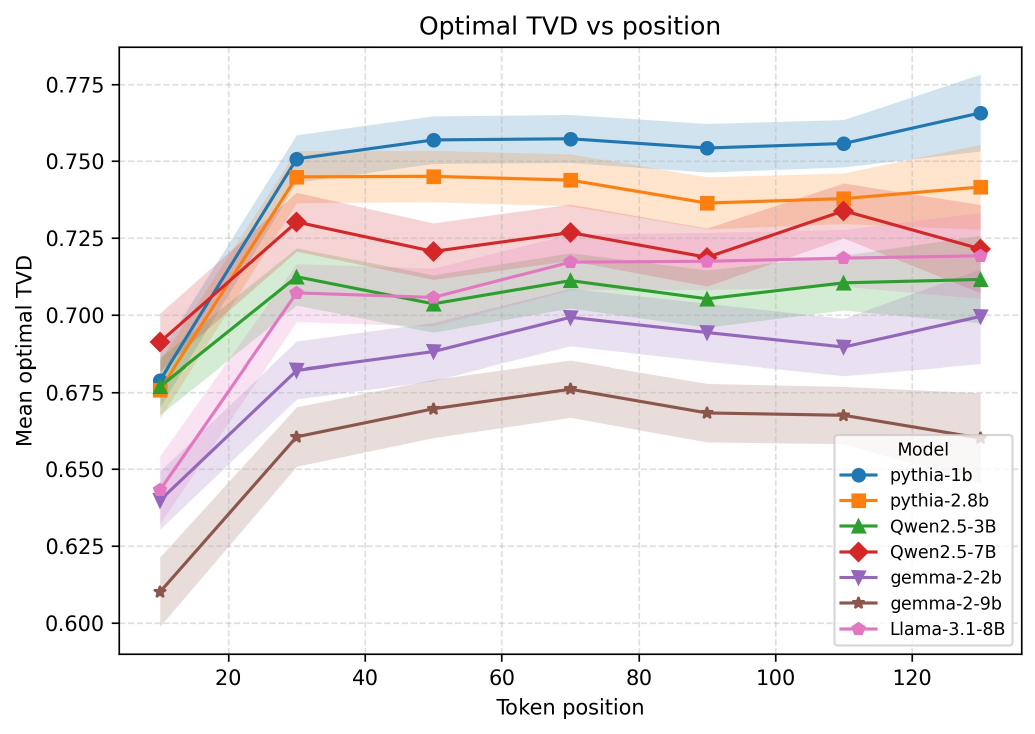}
    \caption{\textbf{Contextual stability of superposition.} Mean Total Variation Distance between the mixed output distribution and the target mixture across token positions. The divergence is slightly lower for the initial tokens ($t < 20$) and then stabilizes for the rest of the context window.}
    \label{fig:tvd_position}
\end{figure}

\section{Layer-wise Linearization Dynamics}
\label{app:layerwise}

To investigate the internal mechanism supporting superposition, we analyze the geometric linearity of transformations layer by layer. Specifically, we measure the extent to which the mapping from the hidden states of layer $\ell$, denoted $\mathbf{H}^{(\ell)}$, to the hidden states of layer $\ell{+}1$, $\mathbf{H}^{(\ell+1)}$, can be approximated by a linear transformation. We use the \textit{linearity score} of~\cite{TransformerIsSecretlyLinear}: given matrices $\mathbf{X}, \mathbf{Y} \in \mathbb{R}^{n \times d}$ obtained by stacking $n$ token feature vectors at layers $\ell$ and $\ell{+}1$, mean-centering, and Frobenius-normalizing them to $\tilde{\mathbf{X}}, \tilde{\mathbf{Y}}$, we set
\[
\mathrm{lin}(\mathbf{X},\mathbf{Y}) \;=\;
1 - \min_{\mathbf{A}\in\mathbb{R}^{d\times d}}
\left\| \tilde{\mathbf{X}}\mathbf{A} - \tilde{\mathbf{Y}} \right\|_F^2,
\]
where the minimizer is the least-squares solution. A score close to $1$ indicates that $\mathbf{H}^{(\ell+1)}$ lies close to an affine reparameterization of $\mathbf{H}^{(\ell)}$.

\begin{figure}[h]
    \centering
    \includegraphics[width=0.7\linewidth]{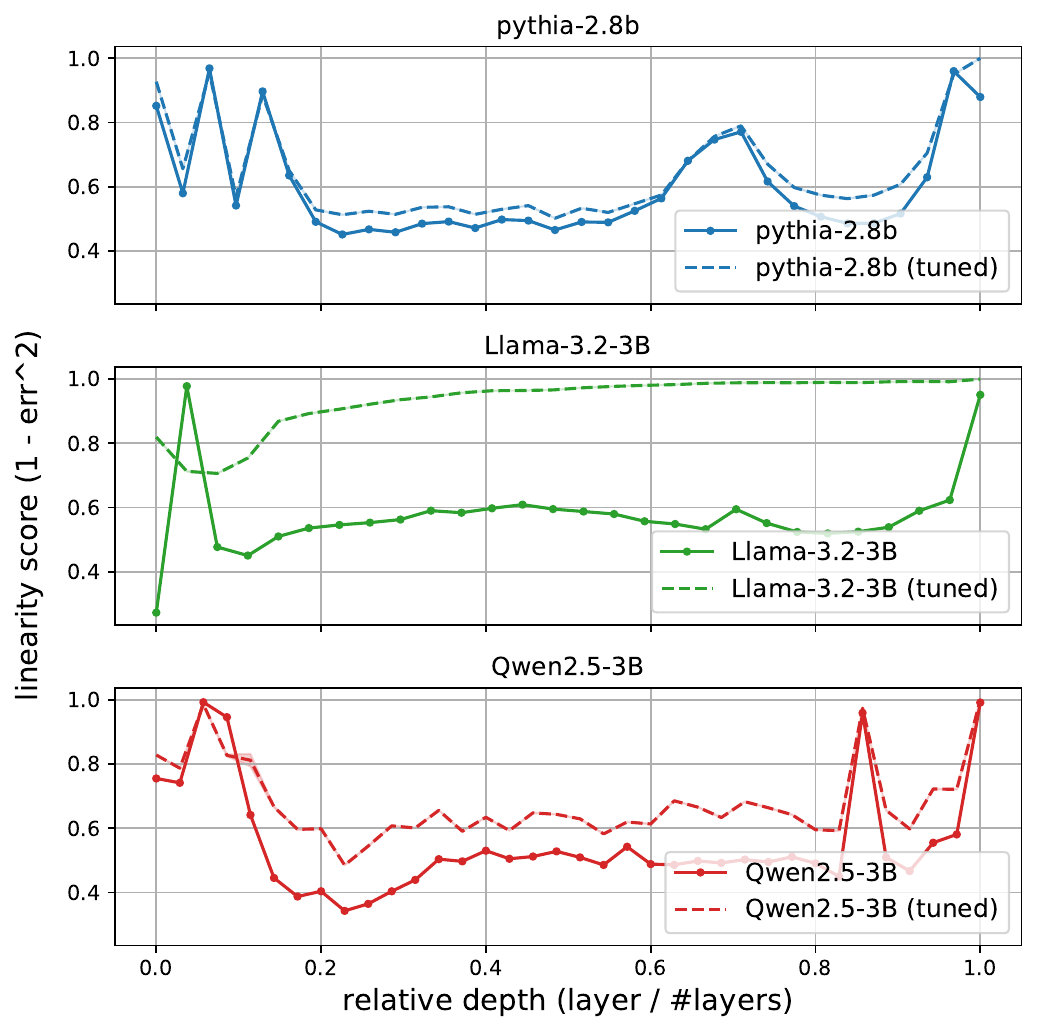}
    \caption{\textbf{Layer-wise linearization dynamics.} Linearity score $1-\mathrm{err}^2$ between consecutive layers for the base and fine-tuned models. The fine-tuned model exhibits consistently higher scores in deep layers.}
    \label{fig:layerwise}
\end{figure}

The depth profile is U-shaped: layers 0--5 are high-linearity (initial embedding integration), the middle layers (6--20) drop to roughly $0.65$, and the final third recovers above $0.95$. This ``terminal linearity'' aligns the high-level features with the unembedding matrix and provides a geometric explanation for the rank preservation results in Sec.~\ref{subsec:rank_analysis}: the quasi-linear final stage prevents the collapse of the superposed signal $z \approx x + y$ before it reaches the output logits. After fine-tuning, the locally modest improvement in linearity scores at each layer is consistent with the substantial reduction in global divergence we observe (Fig.~\ref{fig:rank_curve}, Table~\ref{tab:dist_metrics}), since per-layer non-linear errors compound across depth.

\section{Additional Decoding Variants and Throughput}
\label{app:decoding_extra}
\label{app:throughput}

\subsection{Gradient-Optimized Inference (Logit Arithmetic)}
\label{app:gradient_optimized}

As a lighter-weight alternative to the joint fine-tuning of Sec.~\ref{sec:decoding_obstruction}, we can attempt inference-time separation without altering the large model's weights. We posit that the logits of the target stream $A$ can be recovered via a linear combination of the mixed logits from the large model and the independent logits from a small auxiliary model:
\begin{align}
\tilde{\ell}^{(A)} &= \ell_{\text{large}}(z) + \alpha \cdot \ell_{\text{small}}(A) - \beta \cdot \ell_{\text{small}}(B), \\
\tilde{\ell}^{(B)} &= \ell_{\text{large}}(z) + \alpha \cdot \ell_{\text{small}}(B) - \beta \cdot \ell_{\text{small}}(A).
\end{align}
Keeping the weights of both $M_{\text{large}}$ and $M_{\text{small}}$ frozen, we optimize the scalar coefficients $\alpha, \beta$ using gradient descent to minimize the cross-entropy loss between the reconstructed logits and the ground-truth next tokens on a small calibration dataset. This dynamically balances the large model's expressivity against the small model's directional guidance. However, as shown in Table~\ref{tab:full_quant_metrics}, this purely post-hoc approach achieves lower accuracy than Joint Contrastive decoding.

\subsection{Two-Head Separation via Symmetry Breaking}
\label{app:two_head}

To eliminate the dependency on auxiliary models, we introduce learnable projections $W_A, W_B \in \mathbb{R}^{d \times d}$ and two output heads. The mixed representation becomes $z = \mathrm{Norm}(W_A E(A) + W_B E(B))$, with the projections initialized as $W_k = I + \mathcal{N}(0, \sigma^2)$, $\sigma = 10^{-3}$, $k \in \{A, B\}$, breaking the permutation symmetry of averaging. We train end-to-end with a joint cross-entropy loss over the two heads; naively, this often collapses onto a single stream, so we apply dynamic loss balancing
\[
\mathcal{L} = \alpha_t \mathcal{L}_A + \beta_t \mathcal{L}_B,
\qquad
\alpha_t \propto \Bigl(\tfrac{\bar{\mathcal{L}}_{A,t}}{\bar{\mathcal{L}}_{A,t} + \bar{\mathcal{L}}_{B,t}}\Bigr)^\gamma,
\]
with $\bar{\mathcal{L}}$ an exponential moving average and $\gamma$ a strength hyperparameter. Two-Head and Mixed Distillation results, omitted from the main-paper Table~\ref{tab:quant_metrics}, are reported in Table~\ref{tab:full_quant_metrics} below.

\begin{table}[h]
    \small
    \centering
    \caption{\textbf{Full quantitative recovery and separation metrics} (extension of Table~\ref{tab:quant_metrics}).}
    \label{tab:full_quant_metrics}
    \resizebox{\linewidth}{!}{
    \begin{tabular}{llcccc}
        \toprule
        \textbf{Backbone} & \textbf{Guide} & \textbf{Method} & \textbf{LAMBADA single} & \textbf{LAMBADA superposed} & \textbf{Jaccard} \\
        \midrule
        Qwen2.5-3B   & Qwen2.5-0.5B & Pretrained         & $0.592$ / $0.437$ & $0.168$ & $0.126$ \\
        Qwen2.5-3B   & Qwen2.5-0.5B & Mixed Distillation & $0.592$ / $0.437$ & $0.207$ & $0.133$ \\
        Qwen2.5-3B   & Qwen2.5-0.5B & Joint Contrastive  & $0.592$ / $0.437$ & $\mathbf{0.345}$ & $\mathbf{0.061}$ \\
        \midrule
        Llama-3.2-3B & Llama-3.2-1B & Pretrained        & $0.643$ / $0.540$ & $0.182$ & $0.094$ \\
        Llama-3.2-3B & Llama-3.2-1B & Two Heads          & $0.643$ / $0.540$ & $0.105$ & $0.235$ \\
        Llama-3.2-3B & Llama-3.2-1B & Mixed Distillation & $0.643$ / $0.540$ & $0.174$ & $0.090$ \\
        Llama-3.2-3B & Llama-3.2-1B & Joint Contrastive  & $0.643$ / $0.540$ & $\mathbf{0.430}$ & $\mathbf{0.067}$ \\
        \midrule
        Pythia-6.9B  & Pythia-160m  & Two Heads          & $0.560$ / $0.225$ & $0.080$ & $0.080$ \\
        Pythia-2.8B  & Pythia-160m  & Pretrained         & $0.544$ / $0.225$ & $0.065$ & $0.109$ \\
        Pythia-2.8B  & Pythia-160m  & Mixed Distillation & $0.544$ / $0.225$ & $0.088$ & $0.112$ \\
        Pythia-2.8B  & Pythia-160m  & Two Heads          & $0.544$ / $0.225$ & $0.164$ & $0.117$ \\
        Pythia-1.4B  & Pythia-160m  & Joint Contrastive  & $0.499$ / $0.225$ & $\mathbf{0.110}$ & $\mathbf{0.080}$ \\
        \bottomrule
    \end{tabular}}
\end{table}

\subsection{TinyStories LLM-as-Judge}
\label{app:tinystories}

Following the TinyStories protocol~\cite{eldan2023tinystories} we use an instruction-tuned grader (OpenAI/GPT-5.2) to score completions on \textit{Grammar}, \textit{Creativity}, \textit{Consistency}, and \textit{Plot Coherence} (1--10), plus an estimated writer ``age group''. Independent, raw-superposed, and distilled-with-guided-decoding versions of Pythia-2.8B are compared in Table~\ref{tab:tinystories_qual}.

\begin{table}[h]
    \small
    \centering
    \caption{\textbf{TinyStories LLM-as-judge ratings on Pythia-2.8B}, mean across streams (1--10).}
    \label{tab:tinystories_qual}
    \begin{tabular}{lccc}
        \toprule
        \textbf{Metric} & \textbf{Independent} & \textbf{Superposed} & \textbf{Superposed} \\
                        & \textbf{baseline}    & \textbf{(pretrained)} & \textbf{(fine-tuned)} \\
        \midrule
        Grammar         & 4.52 & 2.59 & \textbf{3.69} \\
        Consistency     & 3.57 & 2.70 & \textbf{3.53} \\
        Creativity      & 5.72 & 2.08 & 2.13 \\
        \bottomrule
    \end{tabular}
\end{table}

\subsection{Single-stream Language Modeling Quality on FineWeb}
\label{app:finetuning_extra}

\begin{table}[h]
    \small
    \centering
    \caption{\textbf{Single-stream LM quality on FineWeb} after the superposition objectives. Joint Contrastive guidance preserves single-stream fluency close to the small-model baseline; Tuned distillation incurs a substantially larger penalty.}
    \label{tab:ppl_nll}
    \begin{tabular}{llcc}
        \toprule
        \textbf{Family} & \textbf{Method} & \textbf{NLL $\downarrow$} & \textbf{PPL $\downarrow$} \\
        \midrule
        \multirow{4}{*}{Qwen2.5}
            & Independent big (3B)        & 2.521 & 12.44 \\
            & Independent small (0.5B)    & 2.942 & 18.96 \\
            & Joint Contrastive (Guided)  & 2.961 & 19.31 \\
            & Tuned distillation          & 4.182 & 65.52 \\
        \midrule
        \multirow{4}{*}{Llama-3.2}
            & Independent big (3B)        & 2.416 & 11.20 \\
            & Independent small (1B)      & 2.605 & 13.53 \\
            & Joint Contrastive (Guided)  & 2.691 & 14.75 \\
            & Tuned distillation          & 4.269 & 71.44 \\
        \midrule
        \multirow{4}{*}{Pythia}
            & Independent big (2.8B)      & 2.645 & 14.09 \\
            & Independent small (160M)    & 3.385 & 29.53 \\
            & Joint Contrastive (Guided)  & 3.392 & 29.72 \\
            & Tuned distillation          & 3.783 & 43.95 \\
        \bottomrule
    \end{tabular}
\end{table}

\subsection{Throughput and Memory}
\label{app:throughput_table}

We benchmark four decoding modes that all consume the same number of input tokens but differ in how they process them: (1) \textbf{Separate} --- fused embeddings $z=\tfrac{1}{2}(E(A)+E(B))$ pass through a single backbone with two output heads; (2) \textbf{Guided} --- fused embeddings through the large backbone plus a small auxiliary model providing per-stream contrastive guidance; (3) \textbf{Big($b{=}2$)} --- vanilla independent inference with batch size 2; (4) \textbf{2$\times$Big(seq)} --- vanilla, the two streams run sequentially. Modes (1)--(2) jointly produce two continuations from a fused hidden state; (3)--(4) process them fully independently and serve as upper- and lower-bound throughput baselines.

\begin{table}[h]
    \small
    \centering
    \caption{\textbf{Generation throughput} (tokens/s, mean $\pm$ 95\% CI over $n{=}100$ runs, prompt length $128$, generation length $128$).}
    \label{tab:throughput}
    \begin{tabular}{lcccc}
        \toprule
        \textbf{Pair} & \textbf{Separate} & \textbf{Guided} & \textbf{Big($b{=}2$)} & \textbf{2$\times$Big(seq)} \\
        \midrule
        Pythia-2.8B + 160M & $109.6 \pm 0.8$ & $79.4 \pm 0.5$ & $113.1 \pm 0.6$ & $56.0 \pm 0.3$ \\
        Pythia-1.4B + 160M & $144.7 \pm 1.0$ & $93.9 \pm 0.6$ & $145.4 \pm 1.0$ & $71.7 \pm 0.5$ \\
        Llama-3B + 1B      & $81.5 \pm 0.6$  & $52.4 \pm 0.3$ & $81.4 \pm 0.5$  & $41.3 \pm 0.2$ \\
        Qwen-3B + 0.5B     & $64.9 \pm 0.5$  & $39.4 \pm 0.2$ & $63.6 \pm 0.3$  & $32.4 \pm 0.2$ \\
        \bottomrule
    \end{tabular}
\end{table}

Separate matches Big($b{=}2$) within $3\%$ across all pairs while producing two coherent continuations from a fused embedding --- a fundamentally different task from independent batch decoding --- and delivers roughly $2\times$ throughput over sequential vanilla decoding. Guided is slower because of the auxiliary forward pass through $M_{\text{small}}$ on each stream but still substantially faster than the sequential baseline. Memory follows the same pattern: Separate adds modest cost over a single vanilla forward pass (e.g., Pythia-2.8B $7.15$ vs.\ $6.34$ GB; Llama-3B $9.26$ vs.\ $6.72$ GB), while Guided incurs higher cost because both models must be resident (e.g., $12.93$ GB peak for Llama-3B+1B).

\section{Attention-Patching: Stratified Examples}
\label{app:patching_examples}

We complement Sec.~\ref{sec:attention_patching} with qualitative examples from three rank buckets, sampled uniformly (we write the leading word-boundary marker as ``\textvisiblespace''). \emph{LOW} ($\mathrm{rank}\in[2,10]$): predominantly punctuation and short function words (\texttt{.}, \texttt{,}, \texttt{\textvisiblespace and}, \texttt{\textvisiblespace in}, \texttt{\textvisiblespace to}, \texttt{\textvisiblespace not}, \texttt{\textvisiblespace are}, \texttt{\textvisiblespace it}, \texttt{\textvisiblespace a}, digits, single-letter BPE fragments; $3/25$ content words). \emph{MID} ($\mathrm{rank}\in[50,500]$): mostly content (\texttt{\textvisiblespace help}, \texttt{\textvisiblespace fresh}, \texttt{\textvisiblespace belief}, \texttt{\textvisiblespace idea}, \texttt{\textvisiblespace installation}, \texttt{\textvisiblespace communities}, \texttt{\textvisiblespace professor}; $\sim\!16/25$ content). \emph{HIGH} ($\mathrm{rank}\in[2000,20000]$): almost entirely content / rare BPE fragments (\texttt{\textvisiblespace recruits}, \texttt{\textvisiblespace bottled}, \texttt{\textvisiblespace Meditation}, \texttt{\textvisiblespace emblem}, \texttt{\textvisiblespace charity}, \texttt{\textvisiblespace widow}, \texttt{\textvisiblespace Immigration}). The function/content split underlying Sec.~\ref{sec:attention_patching} is therefore not an artefact of an ad-hoc heuristic but a systematic property of the rank distribution.

\section{Distributional Metrics for $N=3$ Superposition}
\label{app:n3_metrics}

To evaluate whether the superposition linearity hypothesis holds for more than two streams, we extend our analysis to $N=3$ by mixing the embeddings of three unrelated contexts at context lengths $L=32$ and $L=512$. Table~\ref{tab:n3_dist_metrics} reports the distributional approximation metrics for $N=3$, analogously to the $N=2$ case in Table~\ref{tab:dist_metrics}. Table~\ref{tab:n3_delta} summarizes the degradation in the approximation ratios when transitioning from $N=2$ to $N=3$. While the metrics quantitatively degrade, the approximation ratios remain substantially below 1.0 at both context lengths, confirming that the mixed distribution still approximates the linear mixture of the three independent distributions. The degradation is generally larger at $L=512$ than at $L=32$, indicating the increased difficulty of processing long contexts when more than two streams are mixed.

\begin{table}[h]
    \small
    \centering
    \caption{\textbf{Distributional Approximation Metrics for $N=3$.} Comparison of distances between the model output on mixed inputs vs.\ the target mixture distribution for three streams ($N=3$) at context lengths $L=32$ and $L=512$. Lower is better.}
    \label{tab:n3_dist_metrics}
    \begin{tabular}{l|cc|cc|cc}
        \toprule
        \multirow{2}{*}{\textbf{Model}} & \multicolumn{2}{c|}{\textbf{KL Divergence}} & \multicolumn{2}{c|}{\textbf{JS Divergence}} & \multicolumn{2}{c}{\textbf{Wasserstein}} \\
         & Value & Ratio & Value & Ratio & Value & Ratio \\
        \midrule
        \multicolumn{7}{l}{\emph{Context length $L=32$}} \\
        Pythia-160M & 1.04 & 0.39 & 0.25 & 0.44 & 0.28 & 0.64 \\
        Pythia-410M & 1.73 & 0.44 & 0.39 & 0.54 & 0.31 & 0.71 \\
        Pythia-2.8B & 2.05 & 0.47 & 0.43 & 0.57 & 0.32 & 0.73 \\
        Llama-3.1-8B  & 2.69 & 0.46 & 0.53 & 0.62 & 0.34 & 0.76 \\
        \midrule
        \multicolumn{7}{l}{\emph{Context length $L=512$}} \\
        Pythia-160M & 1.08 & 0.36 & 0.27 & 0.44 & 0.27 & 0.64 \\
        Pythia-410M & 1.82 & 0.42 & 0.42 & 0.56 & 0.31 & 0.73 \\
        Pythia-2.8B & 2.17 & 0.47 & 0.47 & 0.60 & 0.33 & 0.76 \\
        Llama-3.1-8B  & 2.62 & 0.43 & 0.53 & 0.60 & 0.35 & 0.75 \\
        \bottomrule
    \end{tabular}
\end{table}

\begin{table}[h]
    \small
    \centering
    \caption{\textbf{Degradation from $N=2$ to $N=3$.} Change in approximation ratios ($\Delta$) across distances when increasing the number of mixed streams from two to three, at context lengths $L=32$ and $L=512$.}
    \label{tab:n3_delta}
    \begin{tabular}{l|ccc}
        \toprule
        \textbf{Model} & \textbf{KL Ratio} ($\Delta$) & \textbf{JS Ratio} ($\Delta$) & \textbf{WS Ratio} ($\Delta$) \\
        \midrule
        \multicolumn{4}{l}{\emph{Context length $L=32$}} \\
        Pythia-160M & $0.35 \to 0.39$ ($+0.04$) & $0.40 \to 0.44$ ($+0.04$) & $0.63 \to 0.64$ ($+0.01$) \\
        Pythia-410M & $0.39 \to 0.44$ ($+0.05$) & $0.50 \to 0.54$ ($+0.04$) & $0.69 \to 0.71$ ($+0.02$) \\
        Pythia-2.8B & $0.42 \to 0.47$ ($+0.05$) & $0.54 \to 0.57$ ($+0.03$) & $0.71 \to 0.73$ ($+0.02$) \\
        Llama-3.1-8B  & $0.37 \to 0.46$ ($+0.09$) & $0.57 \to 0.62$ ($+0.05$) & $0.69 \to 0.76$ ($+0.07$) \\
        \midrule
        \multicolumn{4}{l}{\emph{Context length $L=512$}} \\
        Pythia-160M & $0.31 \to 0.36$ ($+0.05$) & $0.39 \to 0.44$ ($+0.05$) & $0.61 \to 0.64$ ($+0.03$) \\
        Pythia-410M & $0.35 \to 0.42$ ($+0.07$) & $0.49 \to 0.56$ ($+0.07$) & $0.65 \to 0.73$ ($+0.08$) \\
        Pythia-2.8B & $0.37 \to 0.47$ ($+0.10$) & $0.51 \to 0.60$ ($+0.09$) & $0.69 \to 0.76$ ($+0.07$) \\
        Llama-3.1-8B  & $0.34 \to 0.43$ ($+0.09$) & $0.54 \to 0.60$ ($+0.06$) & $0.67 \to 0.75$ ($+0.08$) \\
        \bottomrule
    \end{tabular}
\end{table}

\section{Hyperparameters and Implementation Details}
\label{app:hparams}

All distillation runs used AdamW with learning rate $10^{-3}$ for the heads and $10^{-4}$ for the backbone, with a ReduceLROnPlateau scheduler (patience $10$, factor $0.5$). Distillation temperature was $2.0$. Batch sizes and gradient accumulation varied with model size: Pythia-2.8B used batch size $8$ with grad accumulation $2$ (effective $16$); Llama-3.2-3B and Qwen2.5-3B used batch size $32$ without accumulation; Pythia-6.9B used batch size $2$ with grad accumulation $4$. All runs used FineWeb or FineWeb-Edu at maximum sequence length $128$.

The guided decoding (Joint Contrastive) models used AdamW, learning rate $5\times 10^{-5}$, effective batch size $16$ ($4 \times 4$ GPUs) for Qwen and Llama, and learning rate $10^{-4}$ with batch size $8$ for Pythia. All runs used FineWeb at maximum sequence length $512$, gradient clipping at $1.0$, $1{,}000$-step linear warmup (Qwen and Llama), and bf16 precision with SDPA attention. The coefficients $\alpha$ and $\beta$ were initialized to $1.0$ and optimized jointly with the backbone. Qwen and Llama checkpoints were taken at step $30{,}000$; the Pythia checkpoint at step $700{,}000$.

For Two-Head separation: dynamic loss-balancing strength $\gamma = 5.0$, EMA momentum $0.99$, warmup $100$ steps, $\alpha_t/\beta_t$ clipped to $[0.1, 10.0]$. ``Norm'' refers to dynamic rescaling that preserves average input norm rather than layer normalization.

\section{LLM as a Judge Prompt}
\label{appendix:llm-as-a-judge-prompt}

\begin{lstlisting}[language={} , caption={The prompt used for evaluating student completions.}, label={lst:llm_judge_prompt}]
The following exercise, the student is given a beginning of a story. The student needs to complete it into a full story. The exercise tests the student's language abilities and creativity. The symbol *** marks the separator between the prescribed beginning and the student's completion:

{} *** {}

Please provide your general assessment about the part written by the student (the one after the *** symbol). Is it grammatically correct? Is it consistent with the beginning of the story? Pay special attention to whether the student manages to complete the sentence which is split in the middle by the separator ***.

Then, grade the student's completion in terms of:
1. Grammar: /10
2. Creativity: /10
3. Consistency: /10
4. Plot coherence: /10

Finally, provide your best guess of what the age of the student might be, as reflected from the completion. Choose from possible age groups: A: 3 or under. B: 4-5. C: 6-7. D: 8-9. E: 10-12. F: 13-16.

Please output your response in the following format:
General assessment: [text]
Grammar: X/10
Creativity: Y/10
Consistency: Z/10
Plot: W/10
Age group: [Letter A/B/C/D/E/F]

For example:
The student's completion of the story is mostly consistent with the beginning of the story. It maintains the focus on Lily and her family, and the sentence split by the separator is completed correctly. However, the student's addition does not fully integrate the shiny decorations found in the attic, which were a significant part of the beginning.
The grammar is generally correct, but there are a few minor errors: <list omitted>.
Overall, the student's completion of the story demonstrates adequate language abilities and creativity, but could benefit from better integration of the shiny decorations and minor grammar improvements.
Grammar: 8/10, Creativity: 7/10, Consistency: 7/10, Age group: E (10-12)
\end{lstlisting}

\section{Impact Statement}
\label{sec:impact_statement}

Our findings can directly improve the scalability and accessibility of large language models by increasing throughput and lowering infrastructure costs. This may benefit applications requiring high-volume or real-time language generation, such as chatbots and large-scale retrieval-augmented systems.

We recognize that parallel processing techniques could also introduce new risks if misapplied, for example by inadvertently mixing unrelated or sensitive streams. Careful validation and monitoring will be important in downstream systems. Overall, this work contributes to the responsible advancement of efficient language modeling and opens new directions for model design that exploit intrinsic architectural properties.


\end{document}